\documentclass[conference]{IEEEtran}
\IEEEoverridecommandlockouts
\usepackage{cite}
\usepackage{amsmath,amssymb,amsfonts}
\usepackage{algorithmic}
\usepackage{graphicx}
\usepackage{textcomp}
\usepackage[ruled,linesnumbered]{algorithm2e}
\usepackage{times}
\usepackage{booktabs, multirow, makecell}
\usepackage{xcolor} 
\def\BibTeX{{\rm B\kern-.05em{\sc i\kern-.025em b}\kern-.08em
    T\kern-.1667em\lower.7ex\hbox{E}\kern-.125emX}}

\begin{document}

\title{MCDGLN: Masked Connection-based Dynamic Graph Learning Network for Autism Spectrum Disorder}

\author{\IEEEauthorblockN{Peng Wang}
\IEEEauthorblockA{\textit{School of Software} \\
\textit{Taiyuan university of Technology}\\
Taiyuan, China \\
beichengnc@gmail.com} \\

\IEEEauthorblockN{Chengxin Gao}
\IEEEauthorblockA{\textit{School of Software} \\
\textit{Taiyuan university of Technology}\\
Taiyuan, China \\
gaochengxin@tyut.edu.cn}

\and

\IEEEauthorblockN{Xin Wen\textsuperscript{*}}
\IEEEauthorblockA{\textit{School of Software} \\
\textit{Taiyuan university of Technology}\\
Taiyuan, China \\
xwen@tyut.edu.cn} \\

\IEEEauthorblockN{Yanrong Hao}
\IEEEauthorblockA{\textit{School of Software} \\
\textit{Taiyuan university of Technology}\\
Taiyuan, China \\
haoyanrong@tyut.edu.cn}

\and

\IEEEauthorblockN{Ruochen Cao}
\IEEEauthorblockA{\textit{School of Software} \\
\textit{Taiyuan university of Technology}\\
Taiyuan, China \\
caoruochen@tyut.edu.cn} \\

\IEEEauthorblockN{Rui Cao\textsuperscript{*}\thanks{* is corresponding author.}}
\IEEEauthorblockA{\textit{School of Software} \\
\textit{Taiyuan University of Technology}\\
Taiyuan, China \\
caorui@tyut.edu.cn} 
}

\maketitle

\begin{abstract}

Autism Spectrum Disorder (ASD) is a neurodevelopmental disorder characterized by complex physiological processes. Previous research has predominantly focused on static cerebral interactions, often neglecting the brain's dynamic nature and the challenges posed by network noise. To address these gaps, we introduce the Masked Connection-based Dynamic Graph Learning Network (MCDGLN). Our approach first segments BOLD signals using sliding temporal windows to capture dynamic brain characteristics. We then employ a specialized weighted edge aggregation (WEA) module, which uses the cross convolution with channel-wise element-wise convolutional kernel, to integrate dynamic functional connectivity and to isolating task-relevant connections. This is followed by topological feature extraction via a hierarchical graph convolutional network (HGCN), with key attributes highlighted by a self-attention module. Crucially, we refine static functional connections using a customized task-specific mask, reducing noise and pruning irrelevant links. The attention-based connection encoder (ACE) then enhances critical connections and compresses static features. The combined features are subsequently used for classification. Applied to the Autism Brain Imaging Data Exchange I (ABIDE I) dataset, our framework achieves a 73.3\% classification accuracy between ASD and Typical Control (TC) groups among 1,035 subjects. The pivotal roles of WEA and ACE in refining connectivity and enhancing classification accuracy underscore their importance in capturing ASD-specific features, offering new insights into the disorder.
\end{abstract}


\section{Introduction}
Autism spectrum disorder (ASD) constitutes a group of widely prevalent neurodevelopmental disorders that alter individuals' autonomous behaviors and social interactions, typically entailing implications that persist throughout a lifetime. Early intervention is pivotal in alleviating the impact of ASD due to the lack of established diagnostic tools or definitive treatments for the disorder\cite{vicedoAutismHeterogeneityHistorical2023,buryBriefReportLearning2022}. Contemporary diagnostic methods primarily hinge on clinicians' subjective evaluations of atypical behavior\cite{hullRestingStateFunctionalConnectivity2017}, a process fraught with biases and a tendency toward underdiagnosing. Consequently, offering fresh insights into ASD is crucial for refining diagnostic processes and enhancing our understanding of the disorder. 

 Neuroimaging has emerged as an essential instrument for investigating cerebral mechanisms, with functional magnetic resonance imaging (fMRI) being a preferred method for identifying patterns and functions in brain networks that help to clarify brain disorders\cite{simpsonAnalyzingComplexFunctional2013}. Traditionally, the technique utilizes individual fMRI data to form maps of the brain functional networks\cite{ahmadiComparativeStudyCorrelation2023}. The process starts with pinpointing specific areas in the brain, followed by the extraction of the fMRI signals that reveal how much oxygen is used in each of these regions. Subsequently, functional connectivity (FC) is evaluated by examining the temporal coherence of neuronal activities across these cerebral regions. Several studies have highlighted notable irregularities in the functional connectivity of individuals with autism spectrum disorders\cite{guoDiagnosingAutismSpectrum2017}, indicating that such abnormal connectivity patterns could aid in diagnosing ASDs. This potential has led to a growing interest in analyzing these connectivity patterns in fMRI data to improve diagnostic criteria for ASD.
 
Over the past several years, the integration of artificial intelligence with medical imaging has received much attention and numerous applications\cite{boscologalazzoExplainableArtificialIntelligence2022}. However, the machine learning methods are somewhat constrained by their dependence on manually extracted features, which may not successfully reveal the deep relation of the data\cite{jieIntegrationTemporalSpatial2018}. Deep learning methods, on the other hand, as the end-to-end framework, have demonstrated strong capabilities in extracting advanced features. \cite{heinsfeldIdentificationAutismSpectrum2018} used an autoencoder to compress input features, which were then treated as input to a classifier and then utilized a 3D convolutional neural network for processing fMRI data.

Graph Neural Network (GNN)\cite{liuSpatialTemporalConvolutionalAttention2022}, a subdivision of deep learning techniques, is renowned for their capacity to accurately delineate topological features\cite{xuHowPowerfulAre2019, velickovicGraphAttentionNetworks2018}. Functional connectivity networks (FCN), which reflect the coordinated response of the cerebral cortex, align well with graph-based data structures, enhancing our grasp of complex brain network patterns and functional connectivity. FCN's inherent compatibility with graph theoretic principles\cite{bullmoreComplexBrainNetworks2009} has spearheaded the integration of GNNs into fMRI research. The commonly utilized static graphs, constructed based on FCNs, provide a snapshot of brain connectivity. Diverse applications, including the use of GNNs for binary classification of brain states\cite{venkatapathyEnsembleGraphNeural2023}, or the innovative work of \cite{azevedoDeepGraphNeural2022}, which merges both time and space dimensions to glean spatial-temporal features, showcase the versatility of GNNs. The BrainGNN model\cite{liBrainGNNInterpretableBrain2021} further tailors the analysis of fMRI data by integrating prior domain knowledge uniquely tailored for each node's transformation matrix. Nonetheless, these static models often overlook the brain's dynamic aspects, an element critical to understanding full brain functionality and highlighted as significant in various studies\cite{mahmoodDeepLearningModel2021}.

Investigating dynamic functional connectivity, researchers seek to establish networks that reflect the brain's fluctuating activities. These dynamic networks are usually derived by partitioning the blood oxygen level-dependent (BOLD) signal into overlapping or distinct windows to form various connectivity matrices\cite{campbellDBGSLDynamicBrain2023}. For instance, the STAGIN model \cite{kimLearningDynamicGraph2021} integrates brain network segments across time frames to create an encompassing embedding. \cite{zhaoSpatialTemporalGraph2022} innovatively applies spatial convolution to individual time points, followed by temporal convolution at each node. \cite{liuBrainTGLDynamicGraph2023} opts for a time series approach to decode patterns from several dynamic graphs, capturing temporal dynamics effectively. However, the limitation of these methods is normally integrating variable networks with the constant static FC, which is often sidelined, potentially resulting in disconnected dynamic mappings. Additionally, conventional functional connectivity typically includes a plethora of irrelevant connections that blur clarity, especially during task execution \cite{boscologalazzoExplainableArtificialIntelligence2022}.

To overcome existing limitations, we introduce a novel masked connectivity-based dynamic graph learning network (MCDGLN) that leverages fMRI data to integrate both static and dynamic functional connectivity for binary classification. We start by partitioning BOLD signals using the sliding window technique and compute dynamic functional connectivity (dFC) via Pearson Correlation Coefficients (PCC). These dFC metrics are then aggregated through stacked weighted edge aggregation (WEA) blocks and a global fusion (GF) module to form task-specific functional connectivity (tsFC), capturing task-relevant features. The And the WEA module includes a cross-convolution with channel-wise element-wise convolutional kernel to extract features.
The network's topological attributes are extracted using a hierarchical graph convolutional network (HGCN) with residual connections, while a self-attention (SA) module highlights key features at the graph level. An attention vector from the SA module assists the attention-based connection encoder (ACE) in compressing static features. We also refine static functional connectivity (sFC) using masked edge drop (MED) to eliminate extraneous features, focusing on essential connections by overlaying sFC onto tsFC masks. The ACE and a multilayer perceptron (MLP) are then used to reduce static feature dimensions and extract relevant information. The final output, combining data from the HGCN, enhances the model's predictive capabilities.
Applied to the ABIDE-I dataset, the model achieved a 73.3\% accuracy rate, demonstrating its effectiveness. 

The primary contributions of this paper are as follows:
\begin{itemize}
    \item We introduce a weighted edge aggregation module, meticulously crafted to amalgamate the dynamic and static attributes of brain networks.
    \item We propose a mask network aiming at pruning redundant connections throughout the brain 
    \item An attention-based mechanism is designed to enhance the key features and to investigate the underlying mechanisms governing the functionality of the brain network.
\end{itemize}

\section{Method}

\subsection{Overview}

\begin{figure*}[t] 
\centering
\includegraphics[width=\textwidth]{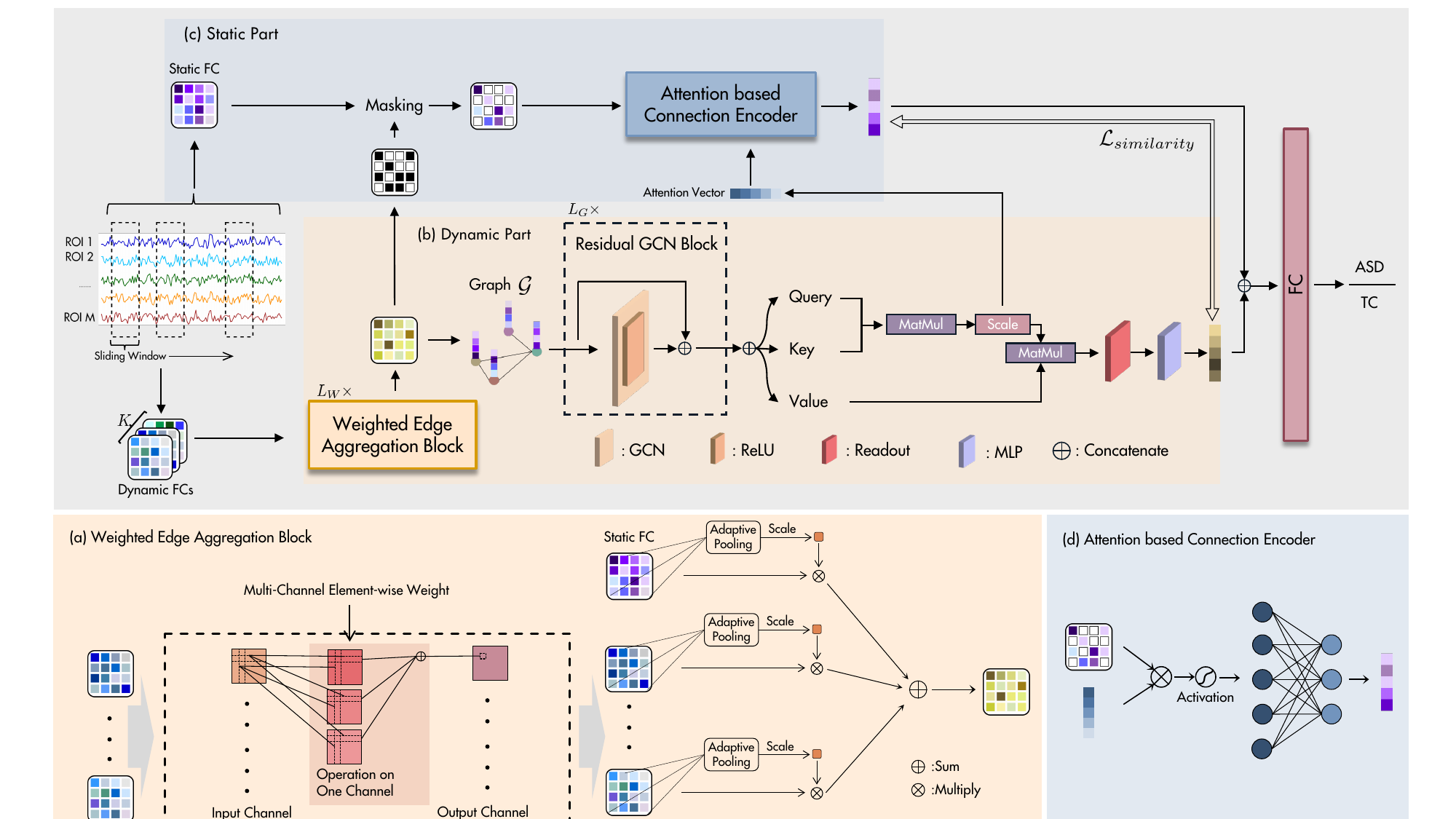}
\caption{The schematic diagram of our proposed framework. The input is the BOLD signal, which is separated by sliding windows to form diversity dFCs for each subject and the output is the prediction of the model. (a) is the section for the process of extracting and compressing the static features and (b) is that of distilling and squeezing the dynamic features.}
\label{fig:model}
\end{figure*}

The comprehensive architecture of MCDGLN introduced in this study is depicted in Fig. \ref{fig:model}. The architecture of the MCDGLN model begins with the input of resting-state fMRI (rs-fMRI) data, from which dynamic functional connectivity (dFC) is computed using pairwise Pearson Correlation Coefficients within sliding windows. These dFCs are aggregated through a weighted edge module to create task-specific functional connectivity (tsFC). A mask generated from tsFC is applied to static functional connectivity (sFC) to refine it. The model then constructs graph data from both sFC and tsFC, using a hierarchical graph convolutional network (HGCN) to extract graph-level embeddings. An attention module highlights key features, and the static and dynamic outputs are combined. The integrated embeddings are finally used for classification predictions.

\subsection{Problem definition}

\subsubsection{The acquisition of dynamic functional connectivity}

In this study, rs-fMRI data are analyzed as multivariate time series. Specially, the variable $x_i\in R^{(M\times T)} (i=1,2,...,N)$ encapsulates the time series of BOLD signals from M distinct ROIs for an individual subject, with T representing the time series length and N denoting the total number of subjects. These time series are segmented using the sliding time window technique, which facilitates the computation of correlation matrices for each segment, hence producing a series of dynamic functional connectivity matrices, denoted as $\mathfrak{c}_{i}$, for every subject. The widely adopted Pearson Correlation Coefficient is employed to quantify the correlations. The Pearson Correlation Coefficient is calculated as follows:

Let $u, v$ be the time series of two ROIs, then the PCC between them is calculated by the following formula:
\begin{equation}
r_{uv}=\frac{\sum\limits_{i=1}^{T}(u_{i}-\bar{u})(v_{i}-\bar{v})}{\sqrt{\sum\limits_{i=1}^{T}(u_{i}-\bar{u})^2}\sqrt{\sum\limits_{i=1}^{T}(v_{i}-\bar{v})^2}}
\label{eq1}
\end{equation}

Where $u_{i},v_{i}$ are the values of variables $u,v$ at the $i$-th time point, and $\bar{u},\bar{v}$ are their respective means. A positive $r_{uv}$ indicates a positive correlation between the two variables, while a negative value indicates a negative correlation. We then set  L as the window size of the sliding window and S as the sliding step of it. Given a subject’s time series $x_{i}$, we can obtain K dynamic functional networks, where K is calculated as: $K=(T-L+S)/S$. The dynamic functional connectivity for each subject is then denoted as: $\mathfrak{c}_{i}^{j}  (i=1,2,…,N;j=1,2,…,K)$. These matrices are then used as input to WEA.

\subsubsection{Graph construction}

A graph data can be represented as $\mathcal{G}=(\mathcal{V},\mathcal{A},\mathcal{X})$ where $\mathcal{V}=\{v:v=1,2,…,M\}$ represents the set of nodes and the adjacency matrix $\mathcal{A}=[a_{uw}:u,w\in \mathcal{V}]\in \{0,1\}^{M\times M}$ denotes whether there is a connection between node u and node w or not. That is, when $a_{uw}=1$, it means there is a connection between the two nodes, and vice versa, when $a_{uw}=0$, it means there is no connection between the two nodes. Then $\mathcal{X}=\{x_v\in \mathbb{R}^{M}:v\in \mathcal{V}\}$ denotes the set of node feature vectors.
Within the realm of constructing graph data utilizing fMRI datasets, the predominant technique involves employing binarized functional connectivity to form the adjacency matrix. 

\subsection{Weighted Edge Aggregation}

\begin{figure}[b] 
\centering
\includegraphics[width=0.8\columnwidth]{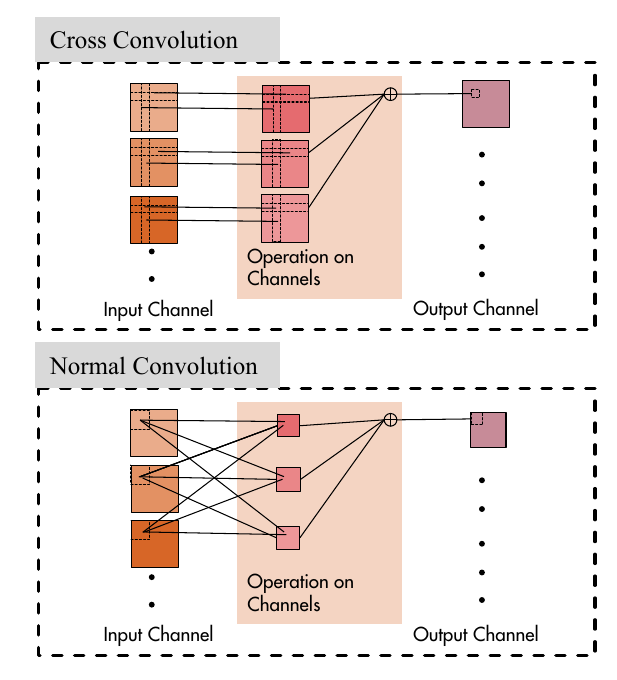}
\caption{The difference between normal convolution and cross convolution. The kernel in the cross convolution is channel-wise and element-wise.}
\label{fig:crosscon}
\end{figure}

The dFC obtained from the sliding time window $\mathfrak{c}_{i} (i=1,2,…,N)$ will be used as an input to WEA. As shown in Fig. \ref{fig:model}(c), the first step is the aggregation of edge information through several cross convolutional layers, which comprised of channel-wise element-wise kernels, and the goal of this process is to fully consider the edge-to-edge relationship. As shown in Fig. \ref{fig:crosscon}, the special feature of the cross convolution operation here is that it applies a channel-wise and element-wise convolutional kernel that captures the dynamic properties in multiple functional connections more efficiently. After updating the FC information through the aggregation process, the global fusion is used to incorporate the static features into the tsFC generation process, to integrate dynamic and static traits into the tsFC, which is the final output of WEA and the input of HGCN.

Specially, suppose $v_{i},v_{j}$ are two ROI nodes with correlation, and the edge connecting them is denoted as $e_{ij}$, then the weighted aggregation process of edges can be expressed as follows:
\begin{equation}
    \hat{e_{ij}}^{l,c}=\sum(e_{im}^{l,c}\cdot w_{im}^{l,c}+e_{nj}^{l,c}\cdot w_{nj}^{l,c})
\label{eq2}
\end{equation}

Where $l=1,2..,L; c=1,2,…,K; n,m=1,2,..,M$ and $M$ represents the total number of nodes, i.e., that of ROIs defined by the parcellation. $l$ is the layer index of the aggregated edge information, which represents the number of times the information is aggregated. $c$ is the index of the channel, where the output channel is kept the same as the input channel. Particularly, the learnable weight matrix $W^{l,c}$ is different in each channel for each layer of the aggregation process, i.e., each edge $e_{ij}^{l,c}$ corresponds to a special learnable weight value $w_{ij}^{l,c}$, which is a learnable parameter of the model. $\hat{e}_{ij}^{l,c}$ are the features of the updated edges after aggregating the information. The task-relevant output is generated by assigning different weight values, deriving from multi-channel element-wise weights to different edges in such a way that the dynamic information is incorporated while considering the relevance of the downstream tasks.

Subsequently, global fusion is implemented to integrate the updated features of dFC and the features of sFC across various channels comprehensively. Specifically, the comprehensive features of functional connectivity undergo condensation into a scalar via a global adaptive pooling block. This scalar then serves as a multiplier for both dFC inputs across multiple channels and sFC. A weighted summation process subsequently leads to the final output, which is encapsulated in the subsequent equation:
\begin{equation}
    \mathfrak{m}_{i}=\sum\limits_{k=1}^{j}\frac{AP(\mathfrak{c}_{i}^{k})}{\|\mathfrak{c}_{i}^{k}\|_{F}}+\frac{AP(\mathfrak{s}_{i})}{\|\mathfrak{s}_{i}\|_{F}}\quad i=1,2,...,N
\end{equation}

Here, $AP$ denotes the adaptive pooling operation. Global fusion facilitates the integration of dynamic features, steered by the static one. The resultant output, tsFC $\mathfrak{m}_{i}$, fulfills a dual role: it serves as the foundation for graph data construction and acts as a mask generator, which generate a mask that is crucial for pruning redundant edges from the static functional connectivity.

\subsection{Masked Edge Drop}

The problem of noise in FC can be very disruptive to the usage of this data for classification task and diagnosis. In this section, a mask that changes with the downstream task is generated based on the tsFC and is applied to remove redundant connections in the sFC to filter out valid connections that are only relevant to the downstream task, i.e., valid edges between ROIs.

Fig. \ref{fig:model}(a) illustrates the process of dropping masked edges. A binary mask matrix, mirroring the dimensions of $\mathfrak{s}_{i}$, is formulated from the tsFC $\mathfrak{m}_i$ that WEA has produced. By calculating the Hadamard product between the mask matrix and $\mathfrak{s}_{i}$, we effectively filter out superfluous connections 
involved in  $\mathfrak{s}_{i}$ while preserving pertinent information.

In calculating the mask matrix according to $\mathfrak{m}_{i}$, the basis is the following formula:
\begin{equation}
f_{m}(\mathfrak{m})=
\begin{cases}
1 & \text{if } \mathfrak{m} \neq 0 \\
0 & \text{if } \mathfrak{m} = 0
\end{cases}
\end{equation}

The output of MED can be formulated as: $\mathfrak{s}\odot f_{m}(\mathfrak{m})$, and it is going to be the input of next part.

\subsection{Hierarchical Graph Convolutional Network}

The graph convolution layer is designed to extract the topological features from the input graph data, denoted as $\mathcal{G}$, aiding in the differentiation of various inputs. In this hierarchical structure, the graph convolutional module processes $\mathcal{G}$ through multiple graph convolutional blocks to capture these features. To address the challenge of over-smoothing, each block incorporates a residual connection. Following this, a self-attention (SA) module highlights important features and generates an attention vector $a_{i}$ that quantifies their significance. This process results in a graph-level feature representation, denoted as $y_{i}^{G}$. The overall workflow is illustrated in Fig. \ref{fig
}(b).

Within our framework, we directly incorporate the graph convolution block (GCN block) as delineated in \cite{kipfSemiSupervisedClassificationGraph2016}. This block iteratively refines node features across successive layers via a message-passing mechanism. For a node v belonging to the vertex set $\mathcal{V}$, with $N(v)$ denoting its neighboring nodes, the message-passing algorithm is mathematically formulated as:
\begin{equation}
    h_{v}^{(k+1)}=\eta_{\Theta}^{(k)}(h_{v}^{(k)}, \gamma_{\Theta}^{(k)}((h_{u}^{(k)}):u\in N_{v}))
\end{equation}

Where $k$ represents the number of layers of the network, $\eta_{\Theta}^{(k)}$ and $\gamma_{\Theta}^{(k)}$ are the trainable functions of the $k$-th layer, which are used to perform the spatial mapping of the vectors. And the convolution principle of GCN written in matrix form can be expressed as follows:
\begin{equation}
    X^{(k+1)}=\hat{D}^{-\frac{1}{2}}\hat{A}\hat{D}^{-\frac{1}{2}}X^{(k)}\Theta^{(k)}
\end{equation}

Where A is the adjacency matrix $\mathcal{A}$, and D represents the diagonal matrix of $\mathcal{A}$, the calculation formula of which is $\hat{D}_{ii}=\sum\limits_{j=0}\hat{\mathcal{A}}_{ij}$.

We define $h_{i}^{(l)}\in \mathbb{R}^{d^{(l)}}$ as the latent features of the $i$-th of the $l$-th block, where $d^{(l)}$ is the dimension of the $l$-th block. The process of joining residuals joining can be formulated as:
\begin{equation}
    H^{(k+1)}=MLP(H^{k}\oplus \sigma_{ReLU}(\hat{D}^{-\frac{1}{2}}\hat{A}\hat{D}^{-\frac{1}{2}}H^{(k)}\Theta^{(k)}))
\end{equation}

Where $k=0,1,2…$ and then use the readout operation to get the graph-level feature embedding of each block. The Readout function used in this paper is the global maximum function, i.e., take the maximum value on each entry. This gives us the graph-level feature embedding of each layer and we use a self-attention mechanism highlights the important features and generate an attention vector $a_{i}$, which can be expressed by the following formula:
\begin{equation}
    y_{i}^{G}=MLP(SA((h^{(1)})\oplus(h^{(2)})...\oplus(h^{(k)})))
\end{equation}

where SA stands for self-attention module. The final output $y_{i}^{G}$ is obtained.

\subsection{Attention-based Connection Encoder}

As shown in Fig. \ref{fig:model}(d), the masked sFC is primarily multiply with the attention vector, to enhance the remaing key connections. Afterwards, a multilayer perceptron is used to compress the features and the output $y_{i}^{M}$ is got in the end, which is going to be integrated with $y_{i}^{G}$ to get the final output. It can be formulated as follows:
\begin{equation}
    y_{i}^{M}=MLP(a_{i}\odot (masked sFC))
\end{equation}
\begin{equation}
    y_{i}=Concat(y_{i}^{M}, y_{i}^{G})
\end{equation}

\subsection{Loss Function}

This thesis performs a binary classification task on ASD and TD using fMRI data. Therefore, the objective function for supervised binary classification is a measure of classification error. We minimize the cross-entropy loss as follows:
\begin{equation}
    \mathcal{L}_c=-\frac{1}{N}\sum\limits_{i=1}^{N}(y_{i}\cdot log(\hat{y}_{i})+(1-y_{i})\cdot log(1-\hat{y}_{i}))
\end{equation}

There is also a loss term that borrows the idea of contrast learning to make the outputs of the hierarchical GCN and the masked connection encoder converge, i.e., align the outputs of the two by adding the contrast regularization $\mathcal{L}_{sim}$, which is calculated as follows:
\begin{equation}
    \mathcal{L}_{sim}=\frac{1}{\mathcal{B}}\sum\limits_{i=1}^{\mathcal{B}}sim(z_{1}^{i}, z_{2}^{i})
\end{equation}

Where $\mathcal{B}$ is the batchsize, the similarity function is a measure of similarity between two vectors, in this paper we use the cosine function of the two vectors to calculate. So, the total loss term loss is:

\begin{equation}
    \mathcal{L}=\mathcal{L}_{c}+\lambda \mathcal{L}_{sim}
\end{equation}

\section{Experiment}

\subsection{Dataset and preprocessing}

In our validation efforts, the framework was applied to the widely recognized Autism Brain Imaging Data Exchange I (ABIDE-I) dataset, a prominent and globally utilized open repository for research on autism spectrum disorder (ASD) and Typically Developing (TD) individuals. ABIDE-I encompasses data from 1112 participants spanning 17 international sites. Profiles for each participant include T1-weighted structural MRI (sMRI), resting-state functional MRI (rsfMRI), and phenotypic information alongside verifiable diagnostic labels. Our study specifically engaged 1035 subjects, consisting of 505 with ASD and 530 TD. To ensure reproducibility and enable equitable comparisons, the investigation leveraged the rsfMRI data subjected to a standardized preprocessing protocol by the Configurable Pipeline for the Analysis of Connectomes (CPAC) via the Preprocessed Connectomes Project (PCP)\cite{cameronNeuroBureauPreprocessing2013}. This preprocessing regimen encompassed steps such as exocranial tissue exclusion, temporal alignment of image slices, motion stabilization, global signal intensity equalization, confounding signal mitigation, and frequency filtering within the range of 0.01-0.1 Hz. Following preprocessing, rsfMRI images were conformed to the MNI152 brain template. The cortical regions of the cerebrum were partitioned into 200 and 116 distinct zones under the Brain Atlas CC200\cite{craddockWholeBrainFMRI2012} and AAL\cite{rollsAutomatedAnatomicalLabelling2020}, from which a region-specific BOLD signal time series was obtained through voxel-wise averaging.

\begin{table*}[t]
    \centering
    \small
    \setlength{\tabcolsep}{5pt} 
    \renewcommand{\arraystretch}{1.2}
    \caption{Performance comparison on CC200 and AAL datasets}
    \label{tab:1}
    \begin{tabular}{@{}lcccccccc@{}}
        \toprule
        \multirow{2}{*}{Method} & \multicolumn{4}{c}{CC200} & \multicolumn{4}{c}{AAL} \\
        \cmidrule(lr){2-5} \cmidrule(l){6-9}
         &  Accuracy & Precision & F1 & AUROC & Accuracy & Precision & F1 & AUROC \\
        \midrule
        vGCN &  $65.7 $ & $67.1 $ & $56.2 $ & $64.7 $ & $66.6 $ & $67.3 $  & $60.0 $ & $65.5 $\\
        GIN & $64.5 $ & $66.2 $ & $58.1 $ & $62.3 $ & $65.1 $ & $65.4 $ & $59.1 $ & $64.9 $ \\ 
        BrainGNN(2021) &  $67.9 $ & $68.8 $ & $31.1 $ & $67.6 $ & $68.0 $ & $68.7 $ & $64.2 $ & $67.0 $ \\
        GATE(2023) &  $70.8 $ & $70.0 $ & $67.3 $ & $69.7 $ & - & -  & - & -\\
        MVS-GCN(2022) & $69.89 $ & - & - & $69.11 $ & $68.92 $ & - & - & $66.44 $ \\
        DGCN(2022) &  $63.7 $ & $55.6 $ & $67.1 $ & $63.6 $ & $68.4 $ & $67.3 $ & $64.6 $ & $67.3 $ \\
        STW-HCN(2024) & - & - & - & - & $70.3 $ & - & - & $71.8 $ \\
        \textbf{Ours} &  $\boldsymbol{73.3 }$ & $\boldsymbol{73.0 }$ & $\boldsymbol{69.7 }$ & $\boldsymbol{71.9 }$ & $\boldsymbol{70.3 }$ & $\boldsymbol{72.0 }$ & $\boldsymbol{66.6 }$ & $69.6$ \\
        \bottomrule
    \end{tabular}
\end{table*}

\subsection{Experimental setup}

\subsubsection{Comparative study}

We compare the proposed MCDGLN with the following eight machine learning methods and deep learning models:

\begin{itemize}
    \item standard GNN: vanilla GCN (v-GCN)\cite{kipfSemiSupervisedClassificationGraph2016}. We built a model of a three-layer GCN, constructing the graph structure in such a way that the PCC matrix is used as the adjacency matrix, and each row corresponding to the ROI is used as the feature embedding of the node. The reported results show the best performance of the model.
    \item GIN\cite{xuHowPowerfulAre2019}. We built a model of a three-layer GIN, and MPL is used as the general mapping function.
    \item BrainGNN\cite{liBrainGNNInterpretableBrain2021}. We used the original implementation provided in using the partial correlation parameter matrix as the adjacency matrix and the PCC matrix as the feature construction graph data of the nodes. The other training parameter settings remained the same as in the original paper.
    \item DGCN\cite{zhaoDynamicGraphConvolutional2022}. We used the original implementation provided in using the functional connectivity matrix as the adjacency matrix, using the PCC matrix as the feature-constructed graph data of the nodes. Other training parameter settings remained consistent with the original paper.
    \item GATE\cite{pengGATEGraphCCA2023}, MVS-GCN\cite{wenMVSGCNPriorBrain2022}, STW-HCN\cite{liuDeepFusionMultitemplate2024}. This paper directly uses the results provided in the original paper.
\end{itemize}

\subsubsection{Ablation study}

In the design of our ablation study, we systematically examine the contribution of each component within our model framework. Specifically, the Weighted Edge Aggregation (WEA), the Attention-based Connection Encoder (ACE), and the Hierarchical Graph Convolutional Network (HGCN) are individually omitted to evaluate their respective roles in the performance of the overall system. Through this methodical approach, we aim to distill the individual impact of each module on the robustness and accuracy of our proposed network architecture.

\begin{figure}[t] 
\centering
\includegraphics[width=\columnwidth]{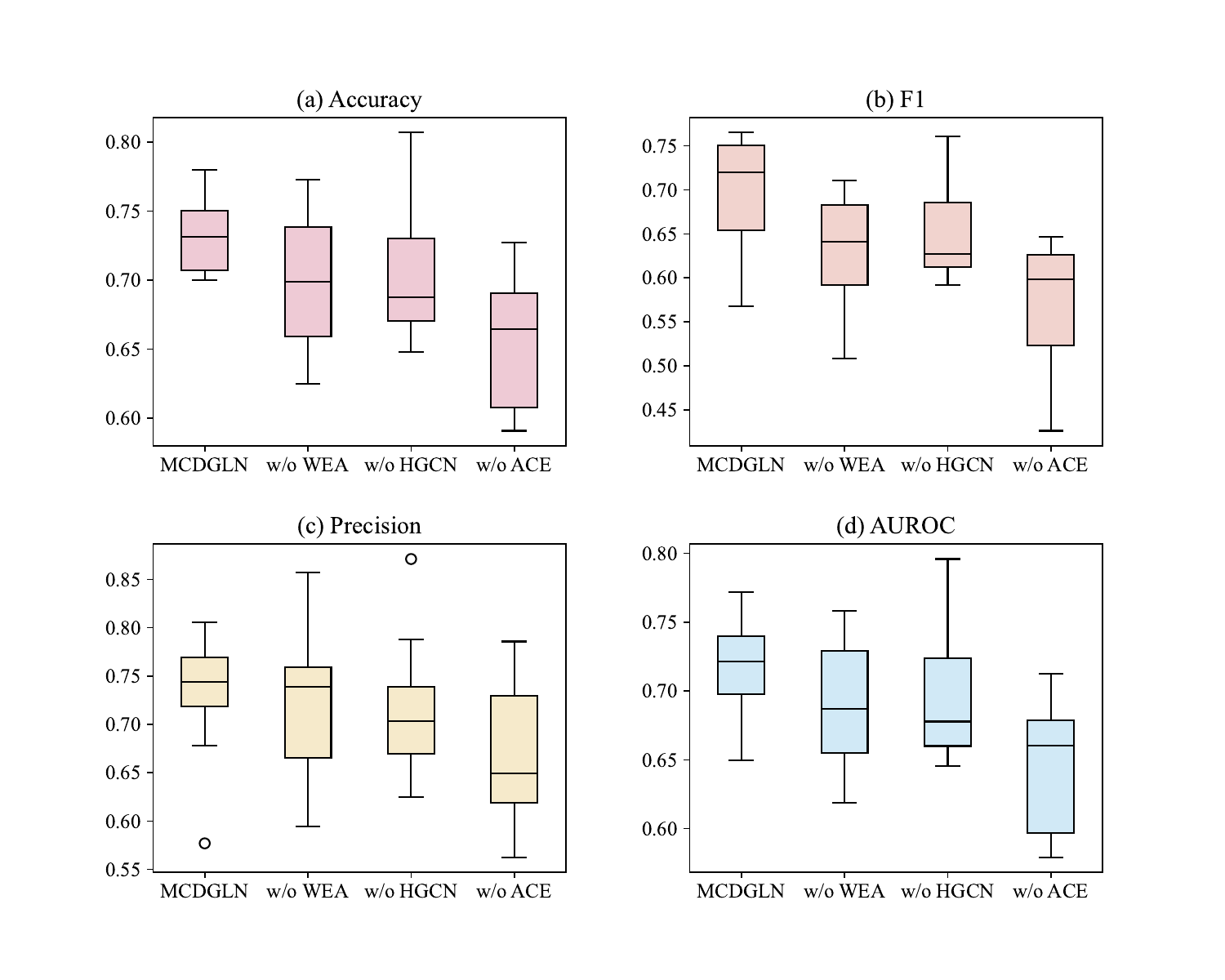}
\caption{The results of ablation study on CC200 atlas}
\label{fig:ablation1}
\end{figure}

\subsection{Metrics}

Our model's performance underwent assessment via a 10-fold cross-validation process. The entirety of the dataset was randomized into 10 equally sized subsamples. In each iteration of the evaluation, one subsample was reserved as the validation set for testing the model, and the remaining nine subsamples were aggregated as the training set. This sequence was executed repeatedly until each subsample had served once as the validation data.

To gauge the efficacy of varying methodologies, we implemented four evaluative metrics: classification accuracy (ACC), precision (PRE), F1 score (F1), and the area under the subject operating characteristic curve (AUC). 

\section{Results and Discussion}

\subsection{Results of comparative study}

The aggregated classification outcomes for each method on our dataset are encapsulated in TABLE \ref{tab:1}. The MCDGLN model introduced in our study registers the highest mean scores across accuracy, AUC, precision, and F1 measures. With an accuracy that exceeds the next-best model, DGCN, by 2\% and outstrips other machine learning techniques by 3\%-6\%, our model demonstrates optimal classification capabilities. In the specific domain of precision, our MCDGLN leads BrainGNN—the second highest performing model—by 3.3\% and surpasses alternative models by a margin of 7\%-11\%, showcasing superior discriminative power for negative samples. On the F1 score front, MCDGLN equates with GATE and consistently outpaces other competitors by 1\%-9\%, reflecting the model's robust performance. For the AUC metric, MCDGLN surpasses the runner-up, DGCN, by 2.6\%, underlining an overall efficient predictive capacity. Notably, our comparative analysis is grounded on the ten-fold cross-validation method, contrary to certain preceding studies\cite{chenAdversarialLearningBased2022, caoUsingDeepGCNIdentify2021} that have only put forth single-partition dataset outcomes, thus not fully capturing model reliability.

In an overall assessment, deep learning frameworks, such as BrainGNN, vGCN, and DGCN, demonstrate superior efficacy over classical machine learning paradigms like SVM across varied evaluative criteria. Specifically, within the cohort of deep learning methodologies, both BrainGNN and DGCN showcase enhanced performance when compared to the static graph convolution approach of vGCN. This underscores the critical impact of dynamic feature capture in the classification process for ASD.

\begin{figure}[t] 
\centering
\includegraphics[width=\columnwidth]{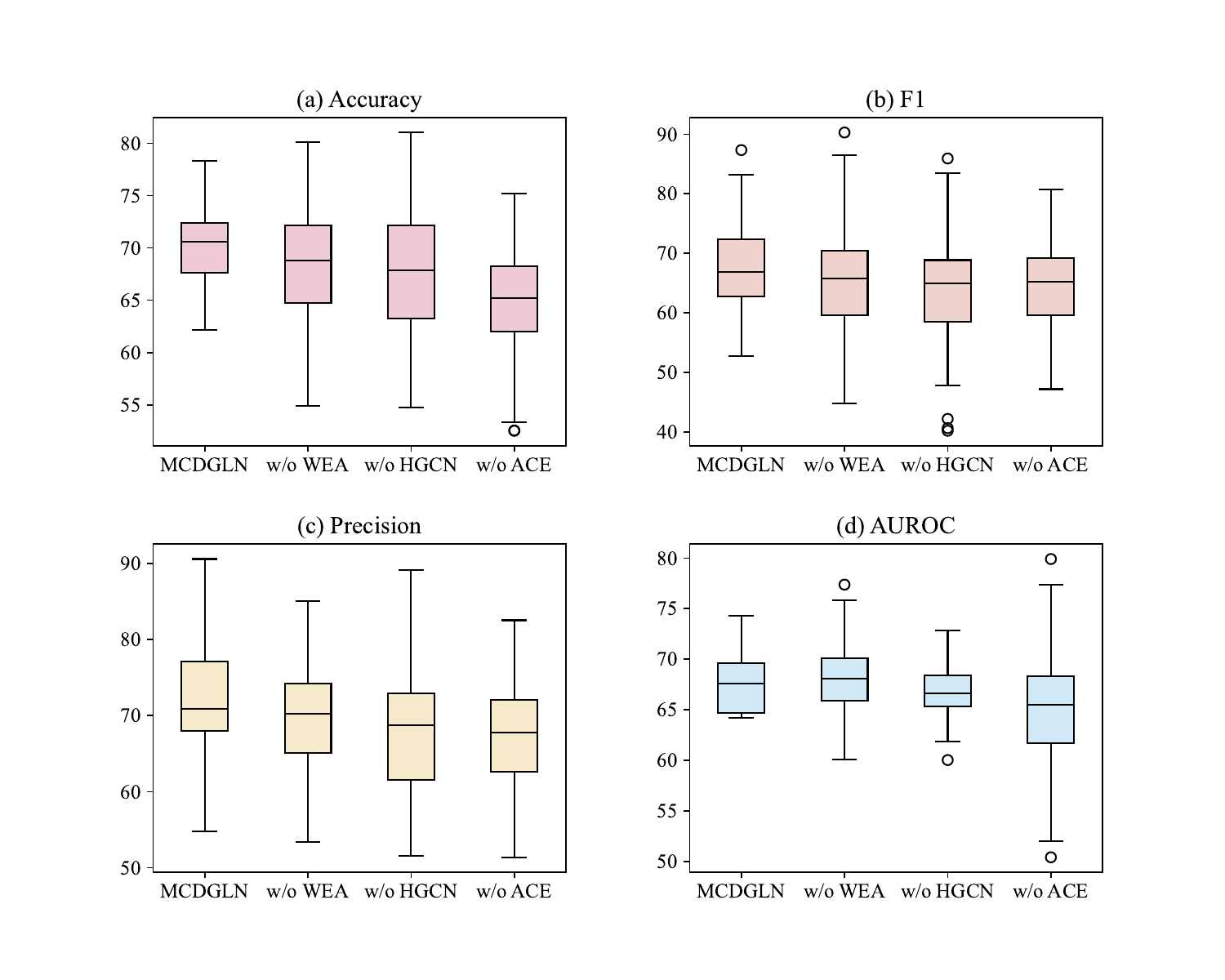}
\caption{The results of ablation study on AAL atlas}
\label{fig:ablation1}
\end{figure}

\subsection{Results of ablation study}

To discern the individual contributions of each module within MCDGLN—namely, the Weighted Edge Aggregation (WEA), the Attention-based Connection Encoder (ACE), and the Hierarchical Graph Convolutional Network (HGCN)—ablation studies were conducted. The outcomes of these studies are detailed in Fig. \ref{fig:ablation1}. The complete, unaltered MCDGLN configuration delivered superior performance across every measured metric, including accuracy, precision, sensitivity, F1 score, and AUC. Notably, the removal of the ACE module resulted in the greatest decrement in metric scores, thereby highlighting its critical role in bolstering the model's efficiency. This can indicate that the masking mechanism and the attention vectors formed based on the attention mechanism suppress the noise in the connection information well and enhance the representation of key features.

\begin{figure}[t] 
\centering
\includegraphics[width=\columnwidth]{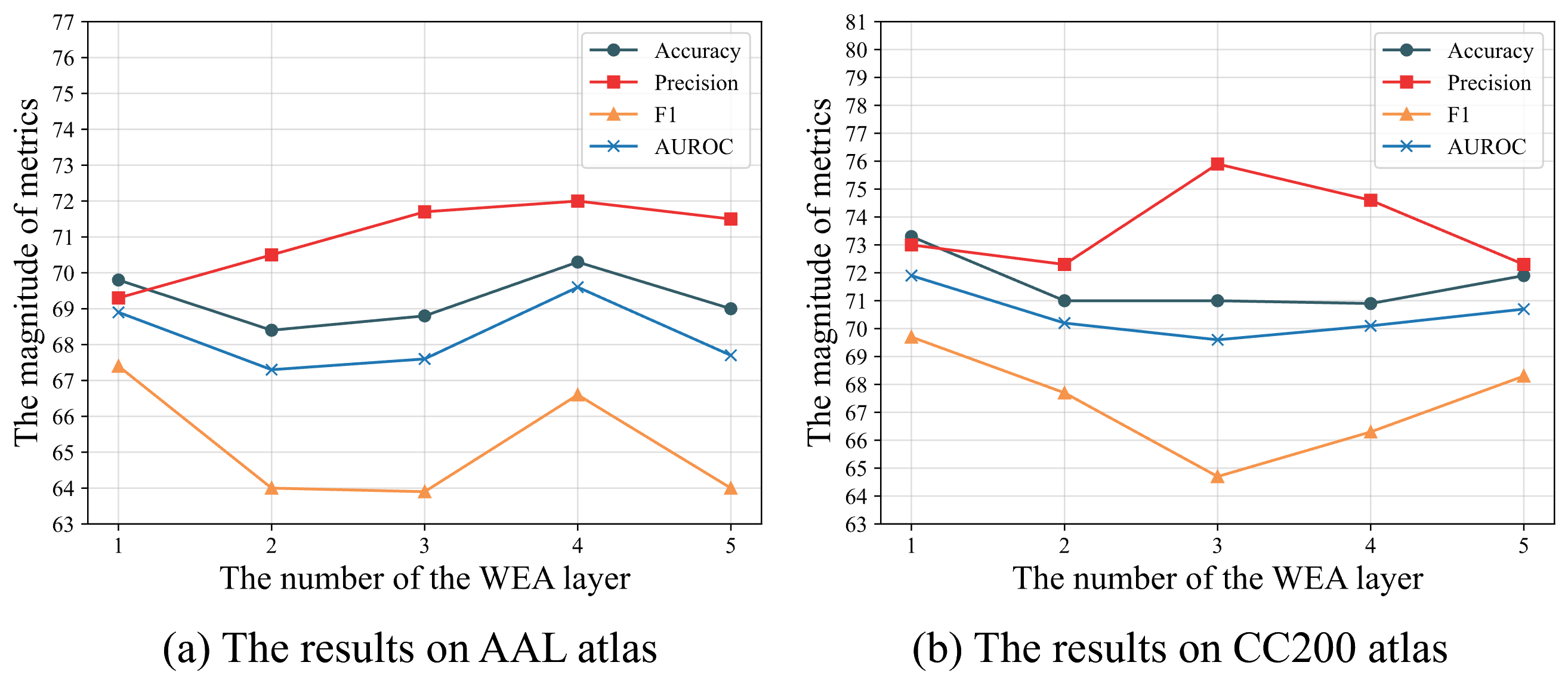}
\caption{The influence of the WEA's layers on two atlases}
\label{fig:ablation3}
\end{figure}

\subsubsection{Impact of WEA's layers}

As shown in Fig. \ref{fig:ablation3}, both panels suggest that the Precision metric is most responsive to the number of WEA layers, peaking at the third layer in both cases. The F1 Score consistently shows a dip at the second WEA layer, indicating potential issues in model performance at this configuration. Accuracy and AUROC metrics remain relatively stable across different WEA layers, highlighting their robustness to changes in the WEA layer configuration.

\subsubsection{Impact of HGCN's layers}

\begin{figure}[t] 
\centering
\includegraphics[width=\columnwidth]{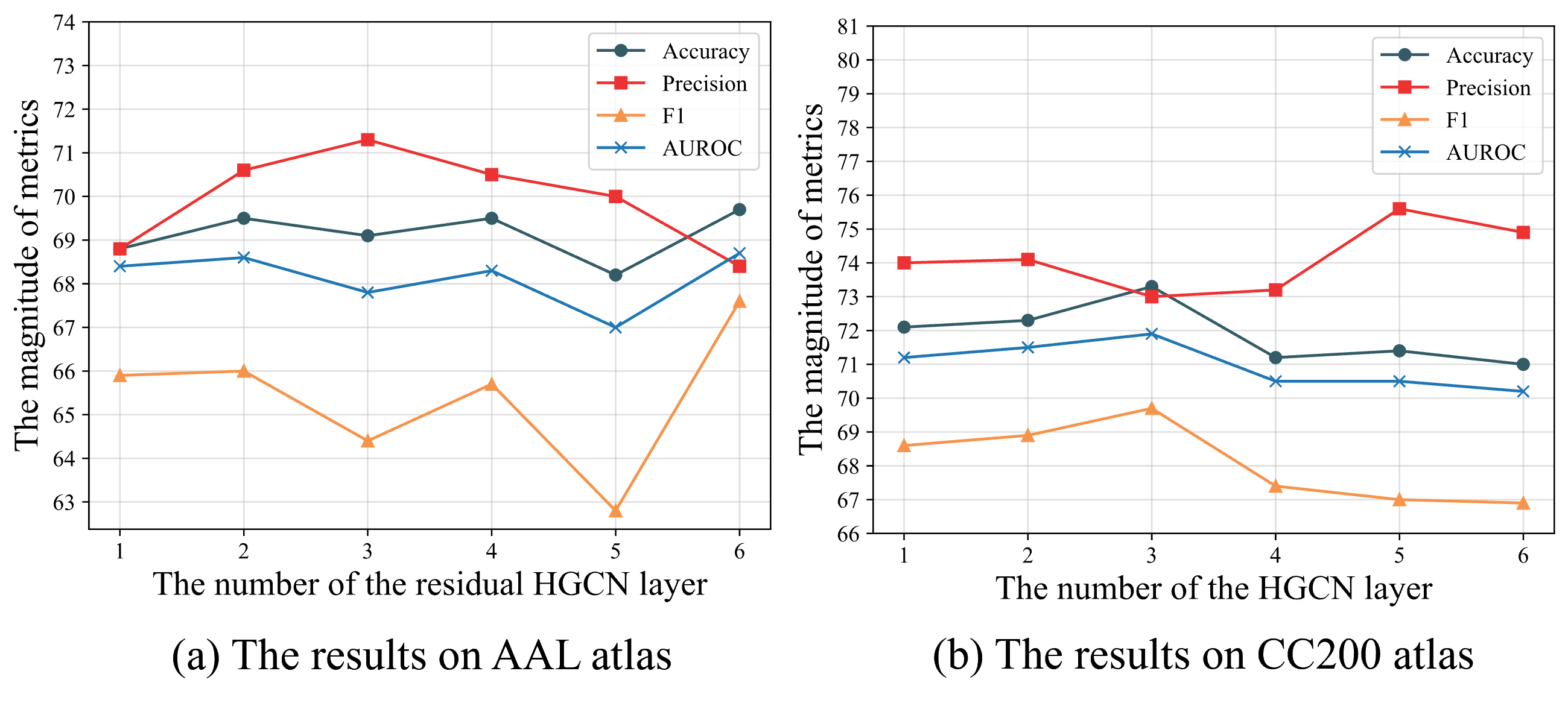}
\caption{The influence of the HGCN's layers on two atlases}
\label{fig:ablation4}
\end{figure}

As shown in Fig. \ref{fig:ablation4}, we can observe trends in performance metrics as the number of layers increases. For instance, in the left plot, both accuracy and AUROC seem to improve gradually with an increasing number of residual HGCN layers; conversely, in the right plot, both "Precision" and "F1" exhibit a rising-falling pattern as the number of HGCN layers varies. Meanwhile, the best number of layers is influenced by the atlas.

\subsection{Functionality of the WEA}

\begin{figure}[t] 
\centering
\includegraphics[width=0.8\columnwidth]{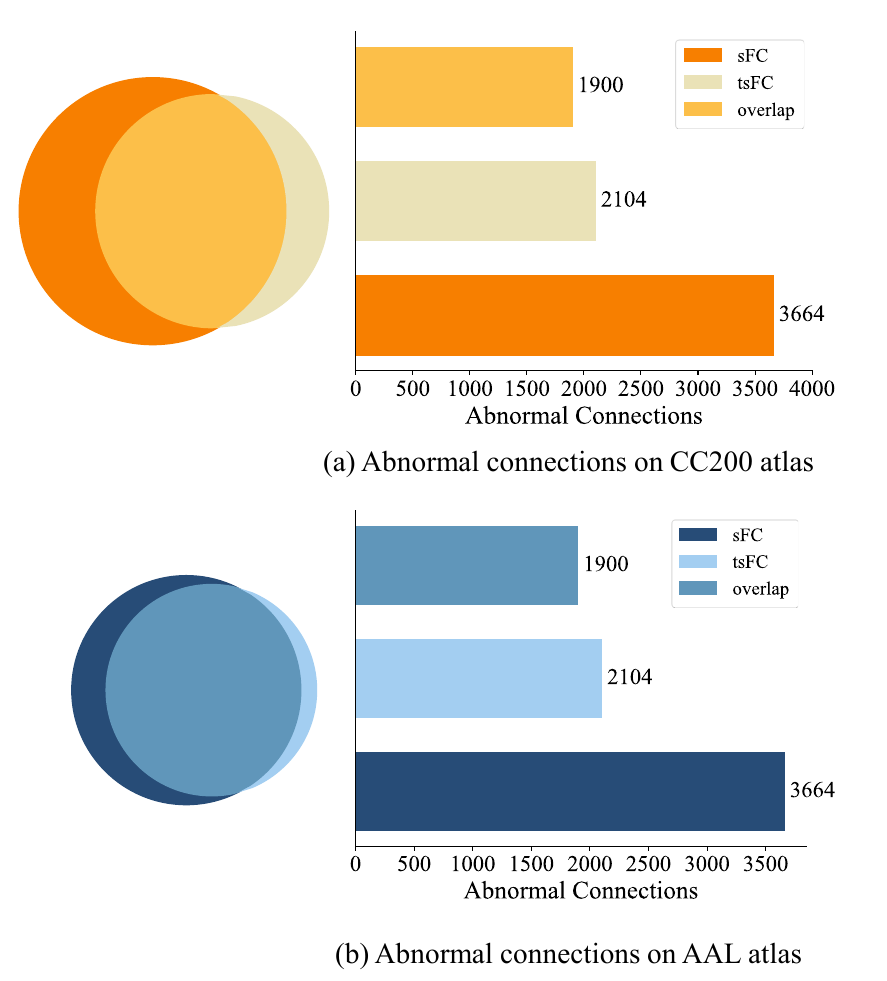}
\caption{The statistics of the abnormal connection found on two atlases}
\label{fig:af}
\end{figure}

This study aims to demonstrate the WEA module's effectiveness and establish its empirical validity by conducting two-sample t-tests on functional connectivity datasets. These tests specifically compare the contrasts in connectivity before and after implementing the WEA module. As illustrated in Fig. \ref{fig:af} and discussed previously, the terms sFC (static functional connectivity) and tsFC (temporal static functional connectivity) describe connectivity patterns, with "overlap" indicating identical connections in both sFC and tsFC. Regardless of the chosen atlas, the majority of abnormal connections identified in sFC are retained in tsFC produced by the WEA module, as evidenced by the extent of overlap. This overlap substantiates the WEA module's efficacy. Furthermore, even without bootstrapping, tsFC created by the WEA module for dynamic feature extraction mirrors sFC characteristics, thereby maintaining the integrity of effective features while also achieving noise reduction.

\section{Conclusion}

In this paper, the MCDGLN model is proposed to extract the dynamic properties of fMRI data thus contributing to the diagnosis of ASD disease. Initially, the model constructs a task-specific functional connectivity network via a weighted edge aggregation. Following this, two critical processes are executed to delineate the network's topological nuances and to eliminate superfluous elements within the original functional connectivity network. The MCDGLN model achieves an accuracy of 73.3\% accuracy, obtaining better results compared to other models. Subsequent experimental results also demonstrate the effectiveness of the proposed module to enhance critical features in functional connectivity and to achieve the goal of noise removal.

\section*{Acknowledgement}

This work was supported by the National Natural Science Foundation of China (62206196), the Natural Science Foundation of Shanxi (202103021223035, 202303021221001) Science and Technology Innovation Program for Higher Education Institutions in Shanxi Province (RD2300004062).

\bibliographystyle{IEEEtran}
\bibliography{IEEEabrv, mybibfile}

\end{document}